\documentclass[sigconf]{acmart}

\usepackage{url}
\usepackage{hyperref}
\usepackage{booktabs}
\usepackage{graphicx}
\usepackage{subfigure}
\usepackage{bm}
\usepackage{color}
\usepackage{amsmath}
\usepackage{amsfonts}
\usepackage{stmaryrd}
\usepackage{stfloats}
\usepackage{xspace}
\usepackage{multirow}
\usepackage{pifont}
\usepackage{pbox}
\usepackage{colortbl}
\usepackage{bbm}
\usepackage{bbding}

\newcommand\ies{\textit{i.e.}}
\newcommand\etal{\textit{et al. }}

\acmSubmissionID{3021}

\begin{document}

\title{Faster Video Moment Retrieval with Point-Level Supervision}


\author{Xun Jiang}
\affiliation{%
  \institution{University of Electronic Science and Technology of China}
  \city{Chengdu}
  \country{China}
}

\author{Zailei Zhou}
\affiliation{%
  \institution{University of Electronic Science and Technology of China}
  \city{Chengdu}
  \country{China}
}

\author{Xing Xu}
\affiliation{%
  \institution{University of Electronic Science and Technology of China}
  \city{Chengdu}
  \country{China}
}

\author{Yang Yang}
\affiliation{%
  \institution{University of Electronic Science and Technology of China}
  \city{Chengdu}
  \country{China}
}

\author{Guoqing Wang}
\affiliation{%
  \institution{University of Electronic Science and Technology of China}
  \city{Chengdu}
  \country{China}
}

\author{Heng Tao Shen}
\affiliation{%
  \institution{University of Electronic Science and Technology of China}
  \city{Chengdu}
  \country{China}
  \\
  \institution{Peng Cheng Laboratory}
  \city{Shenzhen}
  \country{China}
}

\begin{abstract}
  \textit{Video Moment Retrieval} (VMR) aims at retrieving the most relevant events from an untrimmed video with natural language queries. 
  Existing VMR methods suffer from two defects: (1) massive expensive temporal annotations are required to obtain satisfying performance; (2) complicated cross-modal interaction modules are deployed, which lead to high computational cost and low efficiency for the retrieval process. 
  To address these issues, we propose a novel method termed Cheaper and Faster Moment Retrieval (CFMR), which well balances the retrieval accuracy, efficiency, and annotation cost for VMR.
  Specifically, our proposed CFMR method learns from point-level supervision where each annotation is a single frame randomly located within the target moment. 
  It is 6 $\times$ cheaper than the conventional annotations of event boundaries.
  Furthermore, we also design a concept-based multimodal alignment mechanism to bypass the usage of cross-modal interaction modules during the inference process, remarkably improving retrieval efficiency. 
  The experimental results on three widely used VMR benchmarks demonstrate the proposed CFMR method establishes new state-of-the-art with point-level supervision. 
  Moreover, it significantly accelerates the retrieval speed with more than 100 $\times$ FLOPs compared to existing approaches with point-level supervision. 
\end{abstract}

\ccsdesc[500]{Computing methodologies~Activity recognition and understanding}
\ccsdesc[300]{Information systems~Multimedia information systems}

\keywords{Video Moment Retrieval; Point-level Supervised Learning; Model Efficiency; Video Understanding; Multimedia Embedding}

\maketitle

\section{Introduction}
\label{sec:intro}

\begin{figure}[!thb]
	\centering
	\includegraphics[width=\linewidth]{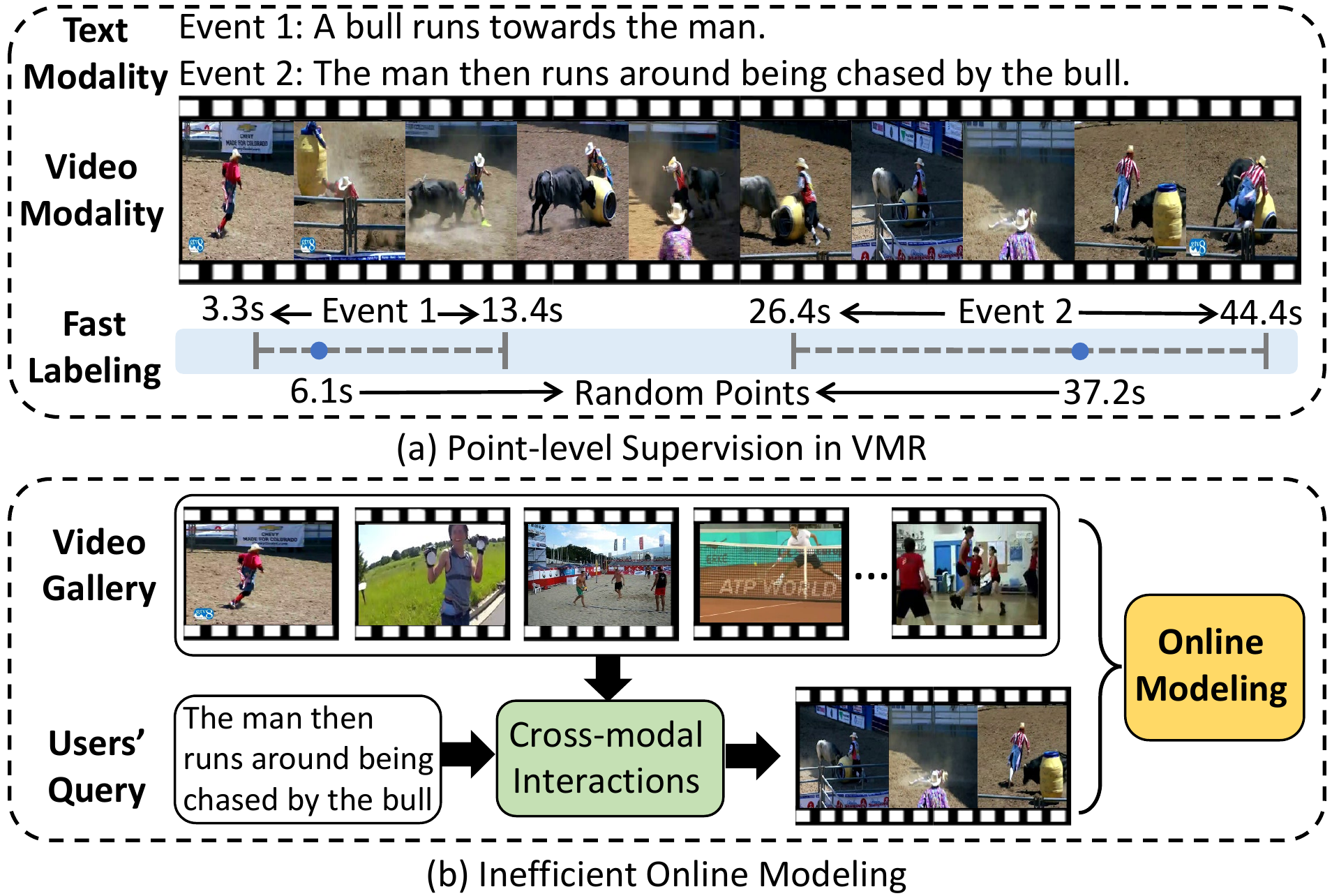}
  
	\caption{
		An illustrative example of PS-VMR and a problem in applications: (a) The annotated points are located within the target moment randomly, which can be labeled much faster and cheaper according to \cite{Fan_SFNet_ECCV2020}. (b) Existing methods have to conduct cross-modal interactions among videos and text queries online, which limits model efficiency heavily.
	}
  
	\label{fig:intro}
\end{figure}

Video Moment Retrieval (VMR) aims at retrieving the startings and endings of particular events in an untrimmed video, with given natural language queries. 
It was first proposed in \cite{gao_TALL_ICCV2017, hendricks_MCN_ICCV2017} and gradually formed two types in terms of the training data annotations: Fully-Supervised VMR (FS-VMR) \cite{gao_TALL_ICCV2017, hendricks_MCN_ICCV2017,mun_LGI_CVPR2020,Wang_CDN_TMM2022,zhang_2DTAN_AAAI2020,Ding_SS_ICCV2021,Gao_FVMR_ICCV2021} and Weakly-Supervised VMR (WS-VMR) \cite{Mithun_TGA_CVPR2019,Lin_SCN_AAAI2020,Zheng_CNM_AAAI2022,Zheng_CPL_CVPR2022,Chen_CWG_AAAI2022}. 
The former, which is the earliest VMR paradigm, contains the conventional VMR methods trained with fully annotated data, where the event boundaries in each video are labeled by annotators deliberately. 
However, it is expensive or even not available to gather clear and determined annotations in most applications, which limits the deployment of these FS-VMR methods. 
To tackle this problem, recently researchers proposed the WS-VMR paradigm trained with video-text matching pairs only, thus significantly reducing the annotation cost.
Unfortunately, as arbitrarily abandoning the temporal supervision signals, the WS-VMR methods obtain much worse performance and generalization compared with the conventional FS-VMR methods, especially on long videos.

To narrow the gap between accuracy and annotation cost, a novel solution for VMR is proposed, \ies, \textit{Point-level Supervised VMR (PS-VMR)} \cite{Cui_ViGA_SIGIR2022, Xu_PSVTG_TMM2022}. 
As shown in Fig. \ref{fig:intro}(a), instead of labeling determined boundaries, the point-level supervision is built on the single frame only, which is randomly located within the target event. 
Compared with fully-supervised labels, point-level labels are much easier to be annotated. 
According to \cite{Fan_SFNet_ECCV2020, Pilhyeon_ACNet_ICCV2021}, point-level supervision is \textit{6 $\times$ cheaper} than traditional ones (the 50s vs. 300s per 1-min video) while labeling the actions or events in an untrimmed video. 
However, all three solutions of VMR employ complicated cross-modal interaction modules to align text and video modalities.
It limits the retrieval process with low efficiency because of the \textit{online modeling}, which can not conduct multimodal learning until receiving users' queries. 
As is depicted in Fig. \ref{fig:intro}(b), existing methods have to repeatedly process the videos with unpredictable queries thus wasting massive computational resources. 
A recent work \cite{Gao_FVMR_ICCV2021} also pointed out this defect and proposed a fast VMR approach to improve retrieval efficiency via a knowledge distillation mechanism. 
Nevertheless, it is highly reliant on expensive annotated data, which still limits the deployment heavily. 
To this end, we are dedicated to exploring a more practical solution for VMR by achieving \textit{balanced performance on accuracy, efficiency, and annotation cost. }

\begin{figure*}[!htb]
	\centering
	\includegraphics[width=0.96\linewidth]{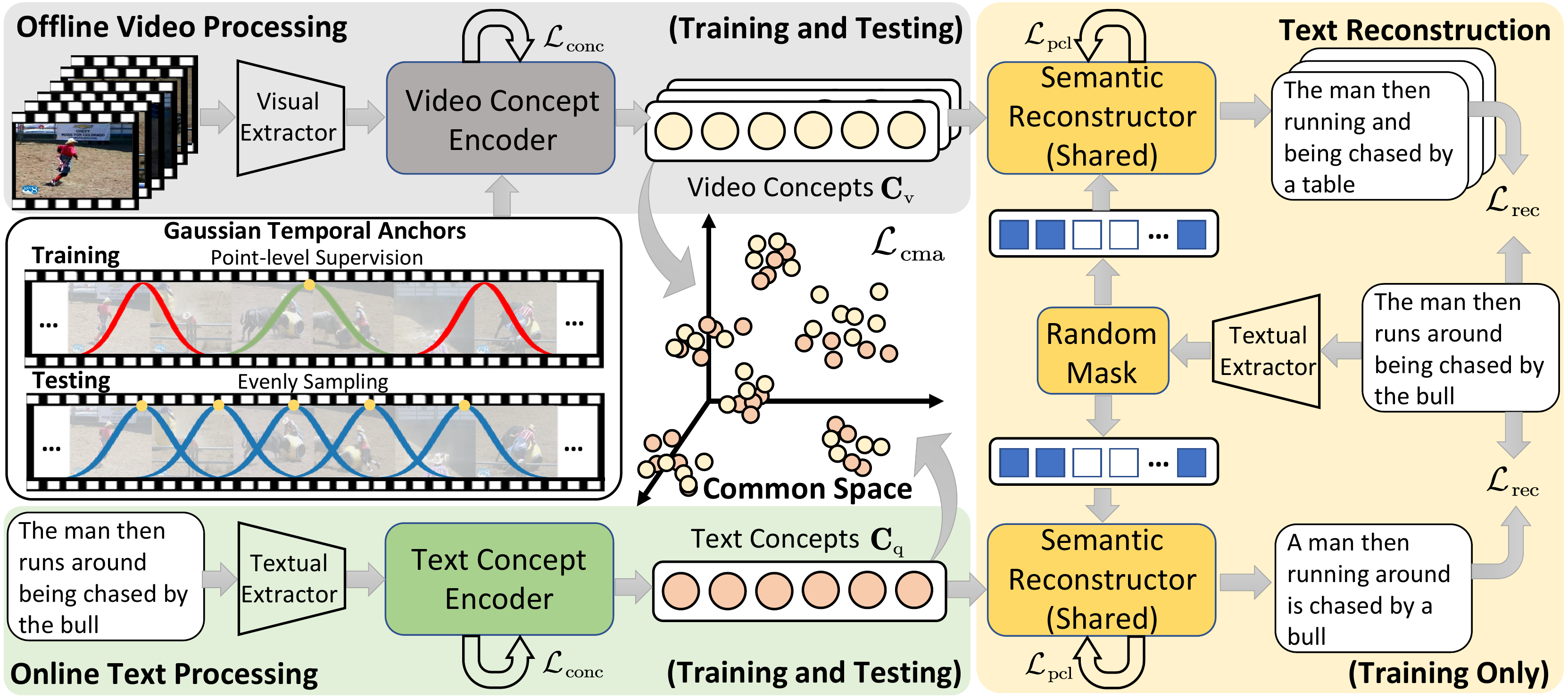}
	\caption{
		Illustration of our proposed CFMR method. 
    The Video Concept Encoder and Text Concept Encoder learn concept representations for the two modalities respectively and project them into a common space to calculate cosine similarity. 
    We also propose the Gaussian temporal anchors, where point-level supervision is adopted during training.
    The Semantic Reconstructor aims to explore the hidden semantics via masked query reconstruction, which is only enabled during training. 
	}
  
	\label{fig:arch}
\end{figure*}

In this paper, we propose a simple but efficient PS-VMR model, termed \textit{Cheaper and Faster Moment Retrieval (CFMR)}, which achieves competitive retrieval performance while significantly reducing annotation cost and computational consumption. 
Specifically, as is depicted in Fig. \ref{fig:arch}, it consists of three Transformer-based components: 
(1) Video Concept Encoder (VCE), (2) Text Concept Encoder (TCE), and (3) Semantic Reconstructor (SR). 
The VCE and TCE modules are designed to encode videos and text queries into a group of concept vectors that represent diverse semantics. 
Furthermore, the SR module is deployed to mine semantic cues for learning concepts of each modality during training. 
In the test stage, we disable the SR module and simply calculate cosine similarity among video and text concepts to retrieve related moments. 
It bypasses online cross-modal interactions thus significantly improving the efficiency and saving massive computational resources.  
We conduct extensive experiments on three widely used VMR benchmarks, \ies, Charades-STA, ActivityNet-Captions, and TACoS. 
The experimental results demonstrate our CFMR method outperforms recent state-of-the-art PS-VMR methods and shows superior comprehensive performance. 

To sum up, our primary contributions are as follows:
\begin{itemize}
  \item We propose a novel Cheaper and Faster Moment Retrieval method that tackles the PS-VMR problem. It is simple but efficient and achieves a fair trade-off among the accuracy, efficiency, annotation cost for VMR. 
  \item We deploy a Point-guided Contrastive Learning module in our model, which effectively utilizes point-level supervision to guide the model to discriminate the semantics hidden in video and text modalities. 
  \item We design a Concept-based Multimodal Alignment module that avoids online cross-modal interactions widely used in previous methods. It allows higher efficiency and saves massive computational resources. 
\end{itemize}

\section{Related Works}
\label{sec:rela}

\noindent
\textbf{Fully-Supervised Video Moment Retrieval.} 
The FS-VMR is the earliest VMR paradigm that proposed by \cite{gao_TALL_ICCV2017, hendricks_MCN_ICCV2017} and have been studied for several years. 
All the data in FS-VMR is fully annotated, where the starting and ending is available for each training instance.
Typically, Zhang ~\etal \cite{zhang_2DTAN_AAAI2020} exploited a two-dimensional temporal adjacent networks to refine the proposal features, then furtherly calculate the correspondence between each moment and text query.
Wang ~\etal \cite{Wang_CDN_TMM2022} designed a dynamic convolution mechanism that fully leveraged the information in the text query to reduce the noise in the video modality. 
Moreover, Mun ~\etal \cite{mun_LGI_CVPR2020} adopted proposal-free framework that need not generate candidates to achieve higher efficiency. 
Gao ~\etal \cite{Gao_FVMR_ICCV2021} adopted a knowledge distillation mechanism to reduce the computational consumption in cross-modal interactions.
Although the FS-VMR methods have achieved remarkable performance, they require massive temporal annotations, which greatly limit their deployment.  

\noindent
\textbf{Weakly-Supervised Video Moment Retrieval.}
The WS-VMR paradigm aims at reducing the heavy annotation cost in the traditional FS-VMR paradigm and attracts many researchers' attention recently. 
It requires the VMR model to learn from fewer temporal annotations \cite{Xun_SVPTR_CVPR2022, Fan_SSL_arxiv2021} or only the matching information among text queries and videos \cite{Mithun_TGA_CVPR2019, Lin_SCN_AAAI2020, Zheng_CNM_AAAI2022, Zheng_CPL_CVPR2022}, or even be trained in a zero-shot manner without any extra supervision \cite{Gao_UVMR_TCSVT2021, Nam_ZSVMR_ICCV2021, Liu_UTVG_AAAI2022}. 
In general, most weakly-supervised learning methods follow \cite{Mithun_TGA_CVPR2019}, which are trained with the matching video-query pairs. 
For example, Lin ~\etal \cite{Lin_SCN_AAAI2020} designed a semantic completion module that predicted the masked important words according to the given visual context for semantic similarity estimation. 
Following \cite{Lin_SCN_AAAI2020}, Zheng ~\etal \cite{Zheng_CNM_AAAI2022,Zheng_CPL_CVPR2022} introduced multiple Gaussian functions into WS-VMR to exploit both positive and negative proposals from the same video. 
However, because of canceling the temporal supervision roughly, the WS-VMR methods perform much worse compared with most FS-VMR methods, especially on long untrimmed videos. 

\noindent
\textbf{Point-level Supervised Learning.}
Point-level Supervised Learning is a weakly-supervised learning paradigm and widely adopted to reduce the annotation cost. 
It turns complex and expensive annotations into rough but extremely simple annotations. 
For example, Bearman ~\etal \cite{Bearman_SSPV_ECCV2016} introduced the point-level supervision by annotating a single pixel for each instance in the semantic segmentation task. 
It is also widely adopted on point-level supervised object detection \cite{PSOD_1, PSOD_2}.
As for video understanding tasks, single-frame supervision is deployed in several recent works \cite{PSVU_1, PSVU_2} to achieve competitive performance on localizing action in untrimmed videos.
Following that, a group of point-level supervised works \cite{Fan_SFNet_ECCV2020, Pilhyeon_ACNet_ICCV2021} on untrimmed videos have been proposed and proves the labeling cost can be reduced 6 times with the point-level annotations. 
Recently, Cui ~\etal \cite{Cui_ViGA_SIGIR2022} and Xu ~\etal \cite{Xu_PSVTG_TMM2022} introduced the point-level supervision signals into VMR, which reduce the heavy annotation cost while achieves competitive retrieval accuracy.

\section{Proposal Method}
\label{sec:prop}

\subsection{Problem Definition}
We represent the training untrimmed video set as $\text{V}$, while the training text query set is represented as $\text{Q}$. 
For each test case, the input can be represented a untrimmed video $v$ and a sentence query $q$. 
Let $t_\text{s}$ and $t_\text{e}$ be the start time and end time of one target video segment respectively, the VMR task can be formulated as $\text{M}_{\theta}\left(v, q ~;\text{V}, \text{Q}, \text{A}\right) \to \{({{t}_\text{s}},{{t}_\text{e}})_{k}\}^\text{K}_{k=1}, ~~~{{t}_\text{s}^{k}}<{{t}_\text{e}^{k}}$.
$({{t}_\text{s}},{{t}_\text{e}})_k$ is the $k$-th retrieved moment in the ranked list of candidates. 
$\text{K}$ is the number of predicted candidates, which is set to 1 in most proposal-free methods. 
The supervision signals are represented as $\text{A}$, which are definite temporal annotations or text-video pairs in FS-VMR and WS-VMR. 
However, in PS-VMR, the supervision signals are rough temporal points randomly located within target events rather than fully annotated labels.

\subsection{Feature Extraction} 
Following the previous methods \cite{Wang_CDN_TMM2022,Zheng_CPL_CVPR2022, Cui_ViGA_SIGIR2022}, we extract the video and text features in an offline manner. 
Firstly, given an untrimmed video ${V}$, we adopt the pre-trained visual backbones \cite{C3D_ICCV2015, I3D_CVPR2017} to extract the visual features. 
Moreover, we employ the GloVe \cite{Glove_EMNLP2014} to extract the word-level embeddings of each sentence query ${Q}$. 
Let $\text{M}_\text{v}\left(\cdot\right)$, $\text{M}_\text{q}\left(\cdot\right)$ be the pre-trained feature extractors of video and text modality respectively, the feature extraction is expressed as follows:
\begin{equation}
      \mathbf{F}_\text{v} = \text{M}_\text{v}\left(V\right) = \{\mathbf{f}_\text{v}^{i}\}_{i=1}^{l_\text{V}}, ~~
      \mathbf{F}_\text{q} = \text{M}_\text{q}\left(Q\right) = \{\mathbf{f}_\text{q}^{j}\}_{j=1}^{l_\text{Q}},
\end{equation}
where the $\mathbf{F}_\text{v} \in \mathbb{R}^{l_\text{V} \times d_v} $ and $\mathbf{F}_\text{q} \in \mathbb{R}^{l_\text{Q} \times d_q} $ represent the extracted video and text features respectively. 
$d_v$ and $d_q$ are feature dimensions for video and text modalities respectively. 
$l_\text{V}$ and $l_\text{Q}$ are the total lengths of video features and number of words in the query.

\subsection{Text/Video Concept Encoder}
As is shown in Fig. \ref{fig:arch}, we adopt two Transformer-based components, \ies, Text Context Encoder (TCE) and Video Context Encoder (VCE), to conduct intra-modality modeling respectively. 
Hence, there are no cross-modal interactions but projecting the two modalities into concept vector sequences with the same lengths.

\noindent
\textbf{Text Modality Encoding.}
Motivated by the success of BERT \cite{Devlin_BERT_NAACL2019}, we employ stacked multiple transformer encoders to learn the global representations for text modality progressively. 
Specifically, given the word-level features, we pad a $\langle\textit{CLS}\rangle$ token at the end of the feature sequence, denoted as $\mathbf{F}_\text{q}^{'} = \{\mathbf{f}_\text{q}^{j}, \mathbf{f}_\text{q}^\text{cls}\}_{j=1}^{l_\text{Q}}$. 
Let $\Phi_\text{q}$ be the text transformer encoders, this procedure can be formulated as:
\begin{equation}
  \mathbf{H}_\text{q}^{l+1} = \Phi_\text{q} \left(f_\text{q}\left(\mathbf{H}_\text{q}^{l}\right)\right), ~~~ \mathbf{H}_\text{q}^{1} = \mathbf{F}_\text{q}^{'},
\end{equation}
where $\mathbf{H}_\text{q}^{l} \in \mathbb{R}^{(l_\text{Q}+1) \times d_h}$ is encoded text feature sequence at $l$-th encoder layer. 
$f_\text{q}\left(\cdot\right)$ is the linear projection layer. 
$d_h$ is the hidden feature dimension. 

\noindent
\textbf{Video Modality Encoding.}
As most videos in the VMR task are untrimmed and contain multiple events and massive irrelevant content, we regard the potential event moments as foregrounds in long untrimmed videos. 
Inspired by the object detection \cite{Redmon_YOLO_CVPR2016,Faster_RCNN_TPAMI2017}, we design the Gaussian temporal anchor mechanism to generate potential foregrounds. 
As demonstrated in Fig. \ref{fig:arch}, all the anchors are represented as Gaussian distributions along the temporal dimension, Furtherly, these Gaussian temporal anchors are used in VCE to re-weight the self-attention matrics, thus enhancing the video content within the corresponding moments in the encoded video features. 
Specifically, given the center $e^{n}$ and width $v^{n}$ of $n$-th Gaussian temporal anchor, we generate the corresponding Gaussian distribution $\mathbf{G}^{n} = \left\{p_{i}^{n}\right\}_{i=1}^{l_\text{V}}$. 
The density of $i$-th video feature $p_{i}^{n}$ can be represented as: 
\begin{equation}
  p_{i}^{n} =\frac{1}{\sqrt{2 \pi}\left({v^{n}} / \gamma\right)} \exp \left(-\frac{\left(i / l_\text{V}-{e^{n}}\right)^2}{2\left(v^{n} / \gamma\right)^2}\right),
\end{equation}
where the $\gamma$ is a scaling hyperparameter. 
Note that $n \in \left\{1,2,\cdots, N\right\}$, $v^{n} = v_\text{max} \times n/N $, where $N$ and $v_\text{max}$ are total number and the max width of anchors for each point. 

By generating adequate Gaussian temporal anchors with different centers and widths, we furtherly employ the VCE to learn the encoded video representation of each corresponding proposal. 
We also adopt the stacked multiple transformer encoders as the fundamental architecture for the VCE and pad a $\langle\textit{CLS}\rangle$ token at the end of video feature sequences, which can be similarly represented as $\mathbf{F}_\text{v}^{'} = \{\mathbf{f}_\text{v}^{i}, \mathbf{f}_\text{v}^\text{cls}\}_{i=1}^{l_\text{V}}$.  
However, the multi-head self-attention weight matrics in each encoder layer will be re-weighted by different Gaussian temporal anchors. 
Let $\Phi_\text{v}$ be the video transformer encoders, the procedure is formulated as follows: 
\begin{equation}
  \mathbf{H}_\text{v n}^{l+1} = \Phi_\text{v} \left(f_\text{v}\left(\mathbf{H}_\text{v n}^{l}\right), \mathbf{G}^{n}\right), ~~~ \mathbf{H}_\text{v n}^{1} = \mathbf{F}_\text{v}^{'},
\end{equation}
where $\mathbf{H}_\text{v n}^{l} \in \mathbb{R}^{(l_\text{V}+1) \times d_h}$ is encoded video feature sequence of $n$-th anchor  at $l$-th encoder layer. 
$f_\text{v}\left(\cdot\right)$ is the linear projection layer. 
Note that the $i$-th row in multi-head self-attention weight matrics in multi-head self-attention of each encoder layer will be re-weighted by $\textbf{G}^n$ through element-wise product.

\noindent
\textbf{Concept Generating.}
We decompose the global representations into multiple concepts for two modalities respectively, where each concept is correlated to a partial semantics and different from others. 
Specifically, given the final encoded global representation of a video proposal $\mathbf{h}_\text{v}^\text{cls}$ and text query $\mathbf{h}_\text{q}^\text{cls}$, we employ two MLPs to project them into high dimensional space and conduct concept decomposing in a multi-head manner.  
Let $\text{Spl}\left(\cdot\right)$ be the dimention split function, this procedure can be formulated as follows: 
\begin{equation}
      \mathbf{C}_\text{v} = \text{Spl}\left(\text{MLP}_\text{v}\left(\mathbf{h}_\text{v}^\text{cls}\right), l_\text{C}\right), ~
      \mathbf{C}_\text{q} = \text{Spl}\left(\text{MLP}_\text{q}\left(\mathbf{h}_\text{q}^\text{cls}\right), l_\text{C}\right),
\end{equation}
where $ \mathbf{C}_\text{v}, \mathbf{C}_\text{q} \in \mathbb{R}^{l_\text{C} \times d_h}$ are text and video concepts respectively. 
$l_\text{C}$ is the total number of concepts.
Moreover, we also adopt the diversity loss \cite{Lin_DivLoss_arXiv2017} that encourages concept vectors to attend to different semantics: 
\begin{equation}
  \mathcal{L}_\text{conc} = \left\|\mathbf{C}_\text{v} \mathbf{C}_\text{v}^{\top}- \mathbf{I}\right\|_F^2 + \left\|\mathbf{C}_\text{q} \mathbf{C}_\text{q}^{\top}- \mathbf{I}\right\|_F^2, 
\end{equation}
where the $\left\|\cdot\right\|$ and $\mathbf{I}$ represent Frobenius norm of a matrix and the identity matrix.

\subsection{Semantic Reconstructor}

\noindent
\textbf{Concept-based Language Reconstruction.}
Inspired by the wide success of self-supervised learning in pre-training models \cite{Devlin_BERT_NAACL2019, Bao_BEIT_arxiv2021}, we adopt the Masked Language Modeling (MLM) to learn the deep semantics from the text queries. 
We randomly mask a part of keywords in the text queries, such as verbs or nouns, and require the SR module to reconstruct each masked word with the given prefix of sentence queries and the generated concepts of two modalities.
Specifically, let $\Psi$ represents stacked multiple transformer decoders, we formulate this procedure as follows: 
\begin{equation}
      \mathbf{P}_\text{v} = \sigma \left(f_\text{r}\left(\Psi\left(\mathbf{C}_\text{v}, \hat{\mathbf{F}}_\text{q}\right)\right)\right), ~~
      \mathbf{P}_\text{q} = \sigma \left(f_\text{r}\left(\Psi\left(\mathbf{C}_\text{q}, \hat{\mathbf{F}}_\text{q}\right)\right)\right),
\end{equation}
where $\sigma\left(\cdot\right)$ and $f_\text{r}\left(\cdot\right)$ represent the softmax function and linear project layer respectively. 
$\hat{\mathbf{F}}_\text{q}$ are masked word-level features, where the masked words are replaced with a particular padding token. 
$\mathbf{P}_\text{v}$ and $\mathbf{P}_\text{q}$ are the predicted probability of each word reconstructed with video and text concepts, $\mathbf{P}_\text{v}, \mathbf{P}_\text{q} \in \mathbb{R}^{(l_\text{Q} \times l_\text{W})}$. 
$l_\text{W}$ is the length of word dictionary.
Moreover, we adopt the log-likelihood loss to estimate the semantic reconstruction: 
\begin{equation}
  \mathcal{L}_{\text{rec}} =-\sum_{j=1}^{l_\text{Q}}\sum_{i=1}^{l_\text{W}} \hat{\mathbf{P}}( \log \mathbf{P}_\text{v}+ \log \mathbf{P}_\text{q} ),
\end{equation}
where $\hat{\mathbf{P}}$ is one-hot vector of the corresponding ground truth. 
The anchor with the smallest reconstruction loss is regarded as the optimal proposal, which is used to calculate the reconstruction loss of video modality. 

\noindent
\textbf{Point-guided Contrastive Learning (PCL).}
Inspired by \cite{Zheng_CNM_AAAI2022, Zheng_CPL_CVPR2022}, we conduct contrastive learning within a group of proposals within the same video to fully utilize the point-level supervision. 
Concretely, we apply the point-level supervision to generate multiple postive proposals. 
We assume that the target event moments are around the annotated points and the points are expected to be the highest density in Gaussian temporal anchors. 
Following the assumption, we generate multiple point-guided anchors during training while the rest parts are sampled as negative anchors, which is depicted in Fig. \ref{fig:arch}. 
Specifically, given three reconstruct loss of the video modality, $\mathcal{L}_{\text{rec}}^{O}$, $\mathcal{L}_{\text{rec}}^{N} $, and $\mathcal{L}_{\text{rec}}^{R}$, which are generated by the optimal anchor, negative anchors, and the complete video, the PCL procedure can be formulated as: 
\begin{equation}
  \begin{aligned}
    \mathcal{L}_\text{pcl}=& \text{Max} \left(\mathcal{L}_{\text{rec}}^{O}-\mathcal{L}_{\text{rec}}^{N}+\alpha_1, 0\right)+\\
                      & \text{Max} \left(\mathcal{L}_{\text{rec}}^{O}-\mathcal{L}_{\text{rec}}^{R}+\alpha_2, 0\right), 
  \end{aligned}
\end{equation}
where $\alpha_1$ and $\alpha_2$ are hyperparameters. 
Note that $\alpha_1$ is expected to higher than $\alpha_2$ as the complete video contains relevant content which is not existed in negative anchors.

\noindent
\textbf{Concept-based Multimodal Alignment (CMA).}
As one of our goals is to give up the clumsy cross-modal interactions, we align the text query with potential video proposals from the perspectives of concepts. 
To this end, we project the concepts of video proposals and their corresponding text queries into a common space, pushing the potential matching pairs closer while the negative pairs further. 
Concretely, similar to PCL module in our model, we regard the optimal anchor and the query as a matching pair while the other as negative pairs, which can be formulated as follows: 
\begin{equation}
  \begin{aligned}
    \mathcal{L}_\text{cma}=& \text{Max} \left(\text{SIM}(\mathbf{C}_{\text{v}}^{N}, \mathbf{C}_\text{q})- \text{SIM}(\mathbf{C}_{\text{v}}^{O}, \mathbf{C}_\text{q}) +\alpha_3, 0\right)+\\
                      & \text{Max} \left(\text{SIM}(\mathbf{C}_{\text{v}}^{R}, \mathbf{C}_\text{q}) - \text{SIM}(\mathbf{C}_{\text{v}}^{O}, \mathbf{C}_\text{q}) +\alpha_4, 0\right) + \\ 
                      & \text{MSE}\left(\mathbf{C}_{\text{v}}^{O}, \mathbf{C}_{\text{q}}\right),
  \end{aligned}
\end{equation}
where $\mathbf{C}_{\text{v}}^{O}$, $\mathbf{C}_{\text{v}}^{N}$, and $\mathbf{C}_{\text{v}}^{R}$ represent the video concepts generated by the optimal video anchor, negative anchors, and the complete video respectively. 
$\mathbf{C}_\text{q}$ is the text concept of the corresponding query. 
$\text{SIM}\left(\cdot\right)$ and $\text{MSE}\left(\cdot\right)$ are the cosine similarity calculation and Mean Squared Error loss. 
$\alpha_3$ and $\alpha_4$ are hyperparameters, similarly, $\alpha_3 \leq \alpha_4$.

\subsection{Training and Inference}
We employ multi-scale strategies in generating Gaussian temporal anchors. 
As illustrated in Fig. \ref{fig:arch}, we take single scale of anchor width as an example here for explicit. 
During training, we regard the point-level supervision as center to generate a postive Gaussian temporal anchors during training. 
The rest of video parts are regarded as two negative anchors. 
Let $\beta_{1}$ and $\beta_{2}$ be balance factors, the overall training objective can be represented as follows:
\begin{equation}
  \mathcal{L}_{\text{total}} = \mathcal{L}_{\text{conc}} + \mathcal{L}_{\text{cma}} + \beta_{1}\mathcal{L}_{\text{rec}} + \beta_{2}\mathcal{L}_{\text{pcl}}.
\end{equation}

During inference, we sample time points along the temporal dimension evenly as the center of Gaussian temporal anchors. 
Moreover, the SR module is disabled and the VCE module can be simply integrated into the video feature extractors, turning untrimmed videos into a group of concepts \textit{in an offline manner}. 
Hence, the online cross-modal interactions are not required. 
We only need to deploy the TCE to learn the text concepts and calculate cosine similarity with the concepts of multiple Gaussian temporal anchors. 
As a result, the computational cost during inference can be significantly reduced in our CFMR method. 

\section{Experiments}

\subsection{Experimental Settings}

\noindent
\textbf{Datasets.}
As we aim to tackle the PS-VMR task, where the temporal annotations are not available during training but only the rough point-level annotations, we adopt the same point-level annotations with \cite{Cui_ViGA_SIGIR2022} for fairness.
The general information of adopted three datasets is summarized as follows:
\textbf{(1) Charades-STA} \cite{gao_TALL_ICCV2017}: 
It contains 6,672 daily life videos in averaged 29.76 seconds and involves 16,128 video-query pairs.
In general, the dataset is split into training and testing parts with 12408 pairs and 3720 pairs, respectively. 
\textbf{(2) TACoS} \cite{regneri_TACoS_TACL2013}: 
It consists of 127 long videos in averaged 287.14 seconds which contain different activities that happened in the kitchen room. 
A standard split \cite{gao_TALL_ICCV2017} consists of 10146, 4589, and 4083 video-sentence pairs for training, validation, and testing, respectively. 
\textbf{(3) ActivityNet-Captions} \cite{ANetCaption_ICCV2017}: 
It contains around 20k open domain videos in averaged 117.61 seconds for video grounding tasks. 
It is split into the training, val, and test of 37421, 17505, and 17031 cases respectively.

\noindent
\textbf{Implementation Details.}
Following the previous methods \cite{zhang_2DTAN_AAAI2020,Cui_ViGA_SIGIR2022,Zheng_CPL_CVPR2022}, we adopt the pre-trained I3D \cite{I3D_CVPR2017} backbone for the Charades-STA dataset and the C3D \cite{C3D_ICCV2015} backbone for the ActivityNet-Captions and TACoS datasets. 
Furthermore, we evenly sample the max lengths of video features $l_\text{V}$ into 200, 200, and 512 while number of concept is set to 7, 8, 3 for the Charades-STA, ActivityNet-Captions, and TACoS datasets respectively. 
The max length of the text query $l_\text{Q}$ is set to 20. 
As for the Gaussian temporal anchors, we set the hyperparameter $\gamma$ to 9. 
The max temporal lengths are set to 0.8, 0.6, 0.3, while the number of center points during inference is set to 8, 4, 15 for the three datasets respectively. 
$v_\text{max}$ is set to 0.55, 0.66, and 0.3 respectively, while $N$ is set to 3. 
We train the CFMR on a single Nvidia RTX 3090 with 32 batchsize. 
The Adam optimizer \cite{Adam_ICLR_Kingma} is employed to update the parameters with the learning rate set to $4 \times 10^{-4}$. 
\textit{For more implementation details, please kindly refer to our supplementary materials and implementation codes.}

\noindent
\textbf{Evaluation Metrics.}
Following the previous methods \cite{zhang_2DTAN_AAAI2020,Zheng_CPL_CVPR2022,Cui_ViGA_SIGIR2022}, we adopt the recall rate as performance metrics, which is denoted as $Recall@ TopK$, $IoU=m$ and abbreviated as $R@K, IoU=m$, where $K$ is the top range number of ranked generated candidates and $m$ is the threshold. 
Specifically, for the three datasets, $k$ is all set to 5, while $m$ is set to $\{0.5, 0.7\}$, $\{0.5, 0.7\}$, and $\{0.3, 0.5\}$ for Charades-STA, ActivityNet-Captions, and TACoS respectively. 
Considering the test time can be influenced by hardwares, we evaluate the floating-point operations per second (FLOPs) and parameters to show the model efficiency.

\subsection{Overall Comparison Results}
\label{sec:exp}
We compare the performance and efficiency with recent state-of-the-art VMR methods, including:
\textbf{(1) FS-VMR methods:} 2DTAN \cite{zhang_2DTAN_AAAI2020}, LGI \cite{mun_LGI_CVPR2020}, SS \cite{Ding_SS_ICCV2021}, FVMR \cite{Gao_FVMR_ICCV2021}, and CDN \cite{Wang_CDN_TMM2022}.
\textbf{(2) WS-VMR methods:} SCN \cite{Lin_SCN_AAAI2020}, CWG \cite{Chen_CWG_AAAI2022}, CNM \cite{Zheng_CNM_AAAI2022}, and CPL \cite{Zheng_CPL_CVPR2022}.
\textbf{(3) PS-VMR methods:} P-LGI \cite{mun_LGI_CVPR2020}, P-SCN \cite{Lin_SCN_AAAI2020}, P-CPL \cite{Zheng_CPL_CVPR2022}, ViGA \cite{Cui_ViGA_SIGIR2022}, and PSVTG \cite{Xu_PSVTG_TMM2022}. 
As the PS-VMR methods are not fully explored yet, we extend \cite{mun_LGI_CVPR2020, Lin_SCN_AAAI2020, Zheng_CPL_CVPR2022} to P-LGI, P-SCN, and P-CPL respectively by introducing the point-level supervision with $L_1$ loss.

\begin{table}[htb!]
  \small
  \centering
  \caption{Accuracy compared with the state-of-the-arts on the Charades-STA dataset using I3D features.}
  
  \resizebox{\linewidth}{!}{
      \begin{tabular}{c|c|cc|cc}
     \toprule
      \multirow{2}{*}{Type} & \multirow{2}{*}{Method} & R@1, & R@1, & R@5, & R@5,  \\ 
      ~ & ~ & IoU=0.5 & IoU=0.7 & IoU=0.5 & IoU=0.7 \\ 
      \hline
      \multirow{5}{*}{FS} & 
      2DTAN \cite{zhang_2DTAN_AAAI2020} & 50.62 & 28.71& 79.92 & 48.52  \\
      ~ & LGI \cite{mun_LGI_CVPR2020} & 59.46 & 35.48 & - & - \\
      ~ & SS \cite{Ding_SS_ICCV2021} & 56.97 & 32.74  & 88.65 & 56.91  \\
      ~ & FVMR \cite{Gao_FVMR_ICCV2021} & 55.01 & 33.74  & 89.17& 57.24      \\
      ~ & CDN \cite{Wang_CDN_TMM2022}  & 51.75 & 29.45 &  78.47 & 54.76     \\
      \hline
      \multirow{4}{*}{ WS} & 
      SCN \cite{Lin_SCN_AAAI2020} & 23.58 & 9.97 & 71.80 & 38.87   \\
      ~ & CNM \cite{Zheng_CNM_AAAI2022} & 35.43 & 15.45  &   -  &   -    \\
      ~ & CWG \cite{Chen_CWG_AAAI2022} & 31.02 & 16.53 & 77.53 & 41.91   \\
      ~ & CPL \cite{Zheng_CPL_CVPR2022} & 49.24 & 22.39 & 84.71 & 52.37   \\
      \hline
      \multirow{6}{*}{PS} & 
      P-LGI \cite{mun_LGI_CVPR2020} & 25.67 & 7.98  &   -  &   -    \\
      ~ & P-SCN \cite{Lin_SCN_AAAI2020}  &   21.73  &   5.52   &   58.57  &   24.97    \\
      ~ & P-CPL \cite{Zheng_CPL_CVPR2022}  &  \textbf{50.27}  & \underline{22.16} & \textbf{84.97} & \underline{52.41} \\
      ~ & ViGA \cite{Cui_ViGA_SIGIR2022}  & 45.05 & 20.27 & 59.87 & 35.24   \\
      ~ & PSVTG \cite{Xu_PSVTG_TMM2022}  & 39.22 & 20.17 & - & -   \\
      \cline{2-6}
      ~ & \textbf{CFMR (Ours)} & \underline{48.14} & \textbf{22.58} & \underline{80.06} & \textbf{56.09} \\ 
      \bottomrule
      \end{tabular}
  }
  \label{tab:charades}

\end{table}

\begin{table}[htb!]
  \small
  \centering
  \caption{Accuracy compared with the state-of-the-arts on the TACoS dataset using C3D features.}
  
  \resizebox{\linewidth}{!}{
  \begin{tabular}{c|c|cc|cc}
  \toprule
  \multirow{2}*{Type} & \multirow{2}*{Method} & R@1, & R@1, & R@5, & R@5,  \\ 
  ~ & ~ & IoU=0.3 & IoU=0.5 & IoU=0.3 & IoU=0.5 \\ 
  \hline
  \multirow{4}{*}{FS} & 
  2DTAN \cite{zhang_2DTAN_AAAI2020} & 37.29 & 25.32 &  57.81  &  45.04  \\
  ~ & SS \cite{Ding_SS_ICCV2021} & 41.33 & 29.56 & 60.65 & 48.01  \\
  ~ & FVMR \cite{Gao_FVMR_ICCV2021} &  41.48  & 29.12 & 64.53 &  50.00 \\
  ~ & CDN \cite{Wang_CDN_TMM2022} & 43.09  & 32.82  & 64.98 &  52.96 \\
  \hline
  \multirow{3}{*}{WS} & 
  SCN \cite{Lin_SCN_AAAI2020} & 11.72 & 4.75 & 23.82 & 11.22  \\
  ~ & CNM \cite{Zheng_CNM_AAAI2022}  & 7.20  & 2.20 &   -  &   -   \\
  ~ & CPL \cite{Zheng_CPL_CVPR2022}  & 11.42 & 4.12 &   33.37  &   15.00   \\
  \hline
  \multirow{5}{*}{PS} & 
  
  P-SCN \cite{Lin_SCN_AAAI2020} &   11.50  & 3.50 &   24.34  &   10.17   \\
  ~ & P-CPL \cite{Zheng_CPL_CVPR2022} &   13.77  & 4.97 &   \underline{34.67}  &   \underline{15.97}   \\
  ~ & ViGA \cite{Cui_ViGA_SIGIR2022} & 19.62 & 8.85  & 26.17  & 14.20 \\
  ~ & PSVTG \cite{Xu_PSVTG_TMM2022}  & \underline{23.64} & \underline{10.00} & - & -   \\
  \cline{2-6}
  ~ & \textbf{CFMR (Ours)}  & \textbf{25.44} & \textbf{12.82}   &   \textbf{49.01}  & \textbf{28.52}   \\ 
  \bottomrule

  \end{tabular}
  }
  
  \label{tab:tacos}
\end{table}

\begin{table}[htb!]
  \small
  \centering
  \caption{Accuracy compared with the state-of-the-arts on the ActivityNet-Captions dataset using C3D features.}
  
  \resizebox{\linewidth}{!}{
      \begin{tabular}{c|c|cc|cc}
     \toprule
      \multirow{2}{*}{Type} & \multirow{2}{*}{Method} & R@1, & R@1, & R@5, & R@5,  \\ 
      ~ & ~ & IoU=0.5 & IoU=0.7 & IoU=0.5 & IoU=0.7 \\ 
      \hline
      \multirow{4}{*}{FS} & 
      2DTAN \cite{zhang_2DTAN_AAAI2020} & 44.51 & 26.54 & 77.13 & 61.96 \\
      ~ & LGI \cite{mun_LGI_CVPR2020}  & 41.51 & 23.07 & - & - \\
      ~ & SS \cite{Ding_SS_ICCV2021} & 46.67 & 27.56 & 78.37 & 63.78 \\
      ~ & FVMR \cite{Gao_FVMR_ICCV2021}  & 45.00 & 26.85 & 77.42& 61.04   \\
      \hline
      \multirow{4}{*}{ WS} & 
      SCN \cite{Lin_SCN_AAAI2020} & 29.22 & - & 55.69  & -  \\
      ~ & CNM \cite{Zheng_CNM_AAAI2022} & 30.26 &   12.81   &   -  &   -    \\
      ~ & CWG \cite{Chen_CWG_AAAI2022} & 29.52 &   -   & 66.61 &   -    \\
      ~ & CPL \cite{Zheng_CPL_CVPR2022} & 31.67 & 13.53 & 43.23 & 22.14   \\
      \hline
      \multirow{6}{*}{PS} & 
      P-LGI \cite{mun_LGI_CVPR2020} &  4.11  &  1.31  &  -   &   -     \\
      ~ & P-SCN \cite{Lin_SCN_AAAI2020} &   21.91  &   12.60    &  49.54  &   21.91    \\
      ~ & P-CPL \cite{Zheng_CPL_CVPR2022}  &   31.10  &   12.79    &  35.37  &   16.61    \\
      ~ & ViGA \cite{Cui_ViGA_SIGIR2022} &  \underline{35.79} & 16.96 & \underline{53.12} & \underline{33.01}   \\
      ~ & PSVTG \cite{Xu_PSVTG_TMM2022} &  35.59 & \textbf{21.98} & - & -   \\
      \cline{2-6}
      ~ & \textbf{CFMR (Ours)} & \textbf{36.97} & \underline{17.33} & \textbf{69.28} & \textbf{49.40}  \\ 
      \bottomrule
      \end{tabular}

  }
  \label{tab:acnet}
  
\end{table}

\noindent
\textbf{Comparisons on Accuracy.}
We compare the performance of our proposed CFMR method with recent state-of-the-art methods using the same video feature extractors. 
The results on the three benchmarks are reported in Table \ref{tab:charades}, \ref{tab:tacos}, and \ref{tab:acnet} respectively, where the best performance and the suboptimal performance are bold and underlined respectively.

\begin{table*}[t!]
  \small
  \centering
  \caption{Overall comparisons on FLOPs ($\times 10^{8}$), Parameters ($\times 10^{6}$), and Accuracy. Considering the length of videos, we select IoU as 0.7, 0.5, 0.3 to evaluate the accuracy for Charades-STA, ActivityNet-Captions, and TACoS respectively. All the results are estimated by averaging the same test cases. For fairness, we ignore offline calculations in compared methods.}
  
  \resizebox{\textwidth}{!}{
      \begin{tabular}{c|c||cccc||cccc||cccc}
      \toprule
      \multirow{2}{*}{Type} & \multirow{2}{*}{Method} & \multicolumn{4}{c||}{Charades-STA} & \multicolumn{4}{c||}{ActivityNet-Captions} & \multicolumn{4}{c}{TACoS} \\ 
      \cline{3-14} 
      ~ & ~ & \text{FLOPs $\downarrow$} & \text{Params $\downarrow$}  & \text{Acc@1 $\uparrow$} & \text{Acc@5 $\uparrow$} & \text{FLOPs $\downarrow$} & \text{Params $\downarrow$} &  \text{Acc@1 $\uparrow$} & \text{Acc@5 $\uparrow$} & \text{FLOPs $\downarrow$} & \text{Params $\downarrow$}  & \text{Acc@1 $\uparrow$} & \text{Acc@5 $\uparrow$}
      \\ 
      \hline
      \multirow{3}{*}{FS} & 
      2D-TAN & 515.90 & 52.43 & 28.71 & 48.52 & 4973.81 & 84.94 & 44.51 & 77.13 & 10615.91 & 52.43  & 37.29 & 57.81 \\
      ~ & LGI    & 34.65 & 29.12  & 35.48 & - & 56.94 & 47.21  & 41.64 & - & - & - & - & - \\
      ~ & CDN    & 136.37 & 238.14  & 29.45 & 54.76 & - & - & - & - & 8594.02 & 238.14 & 43.09  & 64.98\\
      \hline
      \multirow{3}{*}{WS} & 
      SCN      & 8.25 & 5.38 & 9.97 & 38.87 & 11.01 & 7.01  & 29.22 & 55.69 & 19.35 & 6.46 & 11.72 & 23.82\\
      ~ & CNM   & 5.38 & 5.37 & 15.45 & - & 5.33 & 7.01 & 28.70  & - & 15.02 & 6.46 & 7.20 & - \\
      ~ & CPL  & 37.52 & 5.38 & 22.39 & 52.37 & 51.65 & 7.01 & 31.76 & 43.23 & 98.16 & 6.46 & 11.42 & 33.37 \\
      \hline
      \multirow{3}{*}{PS} & 
      P-SCN  & 8.25 & 5.38 & 5.52  & 24.57 & 11.01 & 7.01 & 21.91 & 49.54 & 19.35 & 6.46 & 11.50 & 24.34 \\
      ~ & ViGA     & 16.82 & 12.61 & 19.09 & 32.82 & 17.81 & 12.61 & 35.79 & 53.12 & 38.45 & 12.61 & 19.67  & 26.17\\  
      ~ & \textbf{CFMR (Ours)} & \textbf{0.16} & \textbf{2.98} & \textbf{22.58} & \textbf{56.09} & \textbf{0.22} & \textbf{3.04} & \textbf{36.97} & \textbf{69.28} & \textbf{0.19} & \textbf{2.71} & \textbf{25.44} & \textbf{49.01}\\ 
      \bottomrule
      \end{tabular}
  }
  \label{tab:speed}
  
\end{table*}

Based on the experimental results, we list the following observations: 
(1) Our proposed CFMR method significantly outperforms the recent state-of-the-art PS-VMR methods on most metrics. 
Specifically, it gains at almost 20\% improvement on the metrics of Recall@5 compared with the initial PS-VMR method, \ies, ViGA \cite{Cui_ViGA_SIGIR2022}. 
Compared with the latest PS-VMR method that supports retrieving the single moment only, \ies, PSVTG, the CFMR method with multiple candidates still makes remarkable progress on most metrics and is more practical in retrieval systems . 
It proves the superiority of the proposed CFMR method, which effectively mines hidden semantics with point-level supervision and learns concept-based multimodal alignments only. 
(2) By introducing curriculum learning to explore potential target moment, the latest WS-VMR method, CPL \cite{Zheng_CPL_CVPR2022} and the extended version P-CPL, achieve better performance on Charades-STA dataset on a few low-precision metrics.
However, they show obvious weaknesses in the other two datasets that mainly consist of long untrimmed videos. 
A probable is long untrimmed videos contain more irrelevant content and are more challenging for model robustness. 
It also proves that arbitrarily giving up all temporal supervision to reduce the annotation cost is unreasonable and harmful to the generalization of VMR. 
(3) We also note that our method is suboptimal on the R1@IoU=0.7 than the optimal PSTVG method \cite{Xu_PSVTG_TMM2022} on the ActivityNet-Captions dataset. 
We speculate the reason is there exists a number of recapitulative query cases in this dataset, which are related to the whole video. 
As the proposed CFMR turns the whole video into Gaussian temporal anchors, it is hard to retrieve the complete videos with recapitulative queries.

\noindent
\textbf{Comparisons on Efficiency.} 
To evaluate the comprehensive performance, we count the FLOPs and Parameters of counterpart methods \cite{zhang_2DTAN_AAAI2020,mun_LGI_CVPR2020,Wang_CDN_TMM2022,Lin_SCN_AAAI2020,Zheng_CPL_CVPR2022,Zheng_CNM_AAAI2022,Cui_ViGA_SIGIR2022} based on their official open-source implementation codes. 
Note that the previous FS-VMR method, \ies, FVMR \cite{Gao_FVMR_ICCV2021}, employs knowledge distillation to improve model efficiency. 
Considering the experimental results are measured by test time and the implementation codes are inaccessible, it can hardly make fair comparisons here. 
In theory, we speculate the FVMR should have similar efficiency to ours as the cross-modal interactions can be bypassed in the test stage through knowledge transfer. 
However, the knowledge distillation in FVMR \textit{is highly reliant on fully annotated data}, which still requires massive annotation cost and can not be introduced into the WS-VMR or PS-VMR paradigms.

According to the comparisons listed in Tabel \ref{tab:speed}, we list following observations: 
(1) Our proposed CFMR method is superior to the other methods on efficiency. 
Specifically, compared the previous state-of-the-art PS-VMR method \cite{Cui_ViGA_SIGIR2022}, our method saves more than 100 $\times$ FLOPs with 20\% parameters. 
The reason is our method extracts the video concepts during offline visual feature extraction and models the text query online only. 
Bypasses the complicated cross-modal interactions, it requires much less computational cost. 
(2) The efficiency of our method is less sensitive to video data. 
This is because massive calculation in previous VMR methods can not be propelled until receiving text queries. 
Contrastively, the proposed CFMR process video data in an offline manner thus requires no extra calculation in retrieval but text encoding. 
(3) With higher efficiency, our method still achieves competitive performance on retrieving event moment. 
Our method gets at least 16\% improvements on the Acc@5 than other PS-VMR methods while also reaching the top-rank performance on the Acc@1. 
It explicates that our method effectively maintains retrieval accuracy while significantly reducing computational resources and annotation cost.

\begin{table}[thb!]
	\centering
	\small
	\caption{Ablation study on the Charades-STA dataset.}
 
 \resizebox{\linewidth}{!}{
	\begin{tabular}{cccc|cccccc}
    \toprule
    \multirow{2}*{$\mathcal{L}_\text{conc}$} & \multirow{2}*{$\mathcal{L}_\text{rec}$}  & \multirow{2}*{$\mathcal{L}_\text{pcl}$} & \multirow{2}*{$\mathcal{L}_\text{cma}$} & R@1, & R@1, & R@5, & R@5 \\ 
    ~ & ~ & ~ & ~ & IoU=0.5 & IoU=0.7 & IoU=0.5 & IoU=0.7  \\ 

		\hline 
		\Checkmark   &  \XSolidBrush  & \XSolidBrush   & \XSolidBrush    & 10.94 & 3.74 & 41.81 & 22.49  \\
		\Checkmark   &  \Checkmark & \XSolidBrush   & \XSolidBrush   & 23.44 & 8.03 & 55.39 & 32.27   \\
		\Checkmark   &  \Checkmark & \Checkmark &  \XSolidBrush   & 30.09 & 11.93 & 63.79 & 39.92  \\
		\XSolidBrush   &  \XSolidBrush  & \XSolidBrush   & \Checkmark    & 32.89 & 13.15 & 69.50 & 46.01  \\
		\XSolidBrush   &  \XSolidBrush  & \Checkmark   & \Checkmark    & 35.38 & 15.41 & 71.96 & 43.53  \\
		\XSolidBrush   &  \Checkmark  & \XSolidBrush   & \Checkmark    & 38.17 & 16.00 & 75.24 & 51.16  \\
		\XSolidBrush   &  \Checkmark  & \Checkmark   & \Checkmark    & 47.31 & 21.35 & 79.63 & 54.34  \\
		\Checkmark   & \Checkmark & \Checkmark  & \Checkmark   & \textbf{48.14} & \textbf{22.58} & \textbf{80.06} & \textbf{56.09}   \\
    \bottomrule
  \end{tabular}
  }
	\label{tab:ablation}
 
\end{table}

\subsection{Further Analysis}

\noindent
\textbf{Analysis on Training Objectives.} 
To furtherly explore the effectiveness of each component in our method, we conduct the ablation study on training objectives.  
From the experimental results listed in Table \ref{tab:ablation}, we can observe that: 
(1) By introducing the $\mathcal{L}_\text{rec}$, the proposed CFMR obtains remarkable improvements. 
It demonstrate the SR module is crucial for learning hidden semantics to generate correct concepts for the two modalities. 
(2) The ablated model obtains significantly improvement with $\mathcal{L}_\text{pcl}$. 
The reason is our CFMR method relies on the PCL module to learn semantic cues of matched text and video pairs. 
(3) The CMA module performs an important role in our CFMR method. 
By applying $\mathcal{L}_\text{cma}$, the ablated model achieves great improvements on both two datasets. 
This is because we give up retrieving video moments with cross-modal interactions but aligning the two modality at concept-level. 
Hence, the CMA module is essential for guiding the two encoders to learn common concepts of related video proposals and text queries. 
(4) We can also observe that both $\mathcal{L}_\text{rec}$ and $\mathcal{L}_\text{pcl}$ improve the model performance.
However, the combination of the three loss functions, \ies, $\mathcal{L}_\text{cma}$ + $\mathcal{L}_\text{rec}$ + $\mathcal{L}_\text{pcl}$, significantly boosts the retrieval performance.
The reason is $\mathcal{L}_\text{rec}$ + $\mathcal{L}_\text{pcl}$ can guide our model to learn which video anchor matches the text query best. 
Hence, the CMA process is more effective since the reliability of selected positive video concepts improved.

\begin{table}[thb!]
  \small
  \centering
  \caption{Analysis with respect to concept number $l_\text{C}$ on the Charades-STA dataset with I3D features.}
 
 \resizebox{\linewidth}{!}{
      \begin{tabular}{c|cccc|cc}
     \toprule

      Concept & R@1,    & R@1,    & R@5,    & R@5,    & \multirow{2}{*}{MFLOPs}    & \multirow{2}{*}{Params}  \\ 
      Number & IoU=0.5 & IoU=0.7 & IoU=0.5 & IoU=0.7 & ~ & ~\\ 
      \hline
        4 & 46.47 & 20.96 & 79.58 & 55.20 & 15.74 & 2.78  \\
        5 & 47.09 & 21.42 & 81.06 & 56.44 & 15.81 & 2.85  \\
        6 & 47.36 & 22.52 & 80.06 & 56.41 & 15.87 & 2.91  \\
        7 & 48.14 & 22.58 & 80.06 & 56.09 & 15.94 & 2.98  \\
        8 & 47.58 & 21.85 & 78.85 & 55.79 & 16.01 & 3.04  \\
      \bottomrule
      \end{tabular}
  }
 
 \label{tab:concept_num_charades}
\end{table}
\begin{figure}[thb!]  
 \centering
 \includegraphics[width=\linewidth]{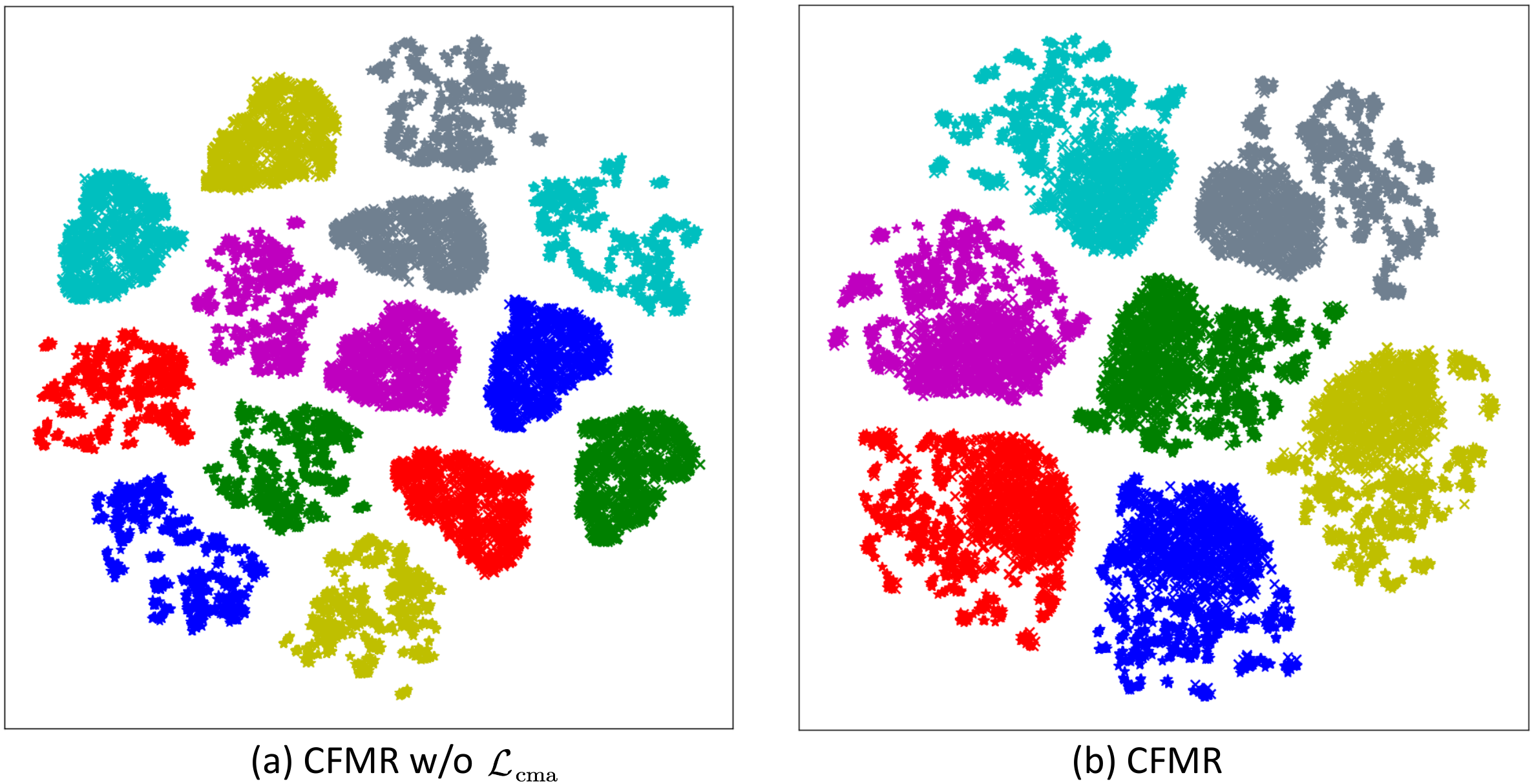}
 \caption{Visualization of multimodal concepts with t-SNE. The colors and shapes represent the numbers and types of modalities respectively, where ``$\star$'' is text concept and ``$\times$'' is video concept.}
 
 \label{fig:concept_clustering}
\end{figure}

\begin{figure}[htb!]  
  \centering
  \includegraphics[width=\linewidth]{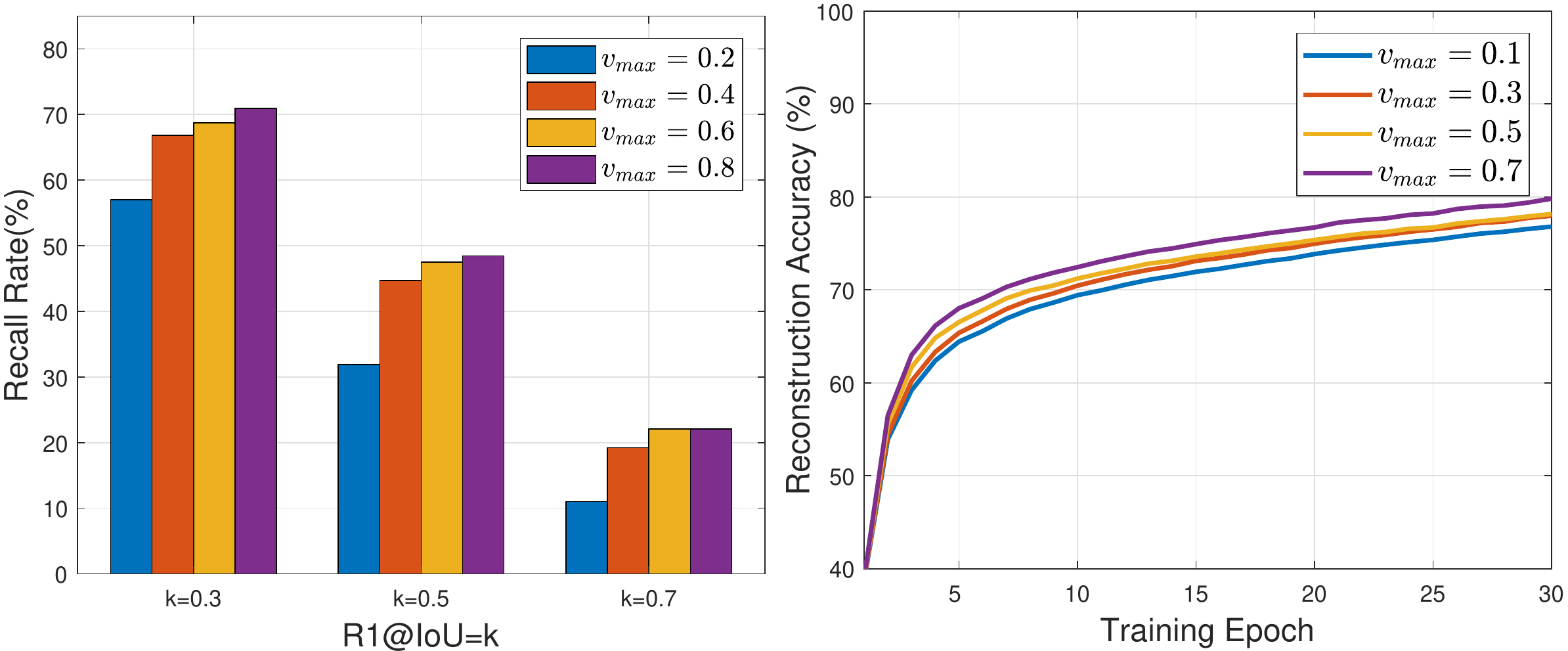}
  \caption{Analysis on performance of moment retrieval and language reconstruction with different $v_\text{max}$ in PCL module.}
 
 \label{fig:PCL}
\end{figure}

\noindent
\textbf{Analysis on Concept-based Multimodal Alignments.}
We also take a closer look at the CMA module procedure to explore how the generated concepts influence the comprehensive performance of the whole model. 
Specifically, we conduct further analysis on the number of concepts $l_\text{C}$ and visualize generated concepts of two modalities via t-SNE \cite{Van_tsne_JMLR2008}. 
By obsering Table \ref{tab:concept_num_charades} and Fig. \ref{fig:concept_clustering}, we can see: 
(1) More concepts will not increase the performance as expected. 
With the number of concept increasing, the recall rate increases slowly but stops at $l_\text{C} = 7$. 
It explicates that information granularity has impacts on learning the semantics in two modalities.
Hence, an appropriate setting is critical in estimating the semantic similarity. 
(2) The computational consumption increases linearly with more concepts but is still superior to the other methods. 
This is because our CFMR methods tackle the VMR task without heavy cross-modal interactions in the inference stage, which means the video can be processed in an offline manner with visual extractors. 
(3) By applying $\mathcal{L}_\text{cma}$, the concepts show a clustering tendency and the concepts of video proposals are hybrid with the concepts of corresponding text queries. 
It demonstrates that the CMA module in our method effectively learns diverse semantics and generates similar concepts for related video proposals and text queries.

\begin{figure}[htb!]  
  \centering
  \includegraphics[width=\linewidth]{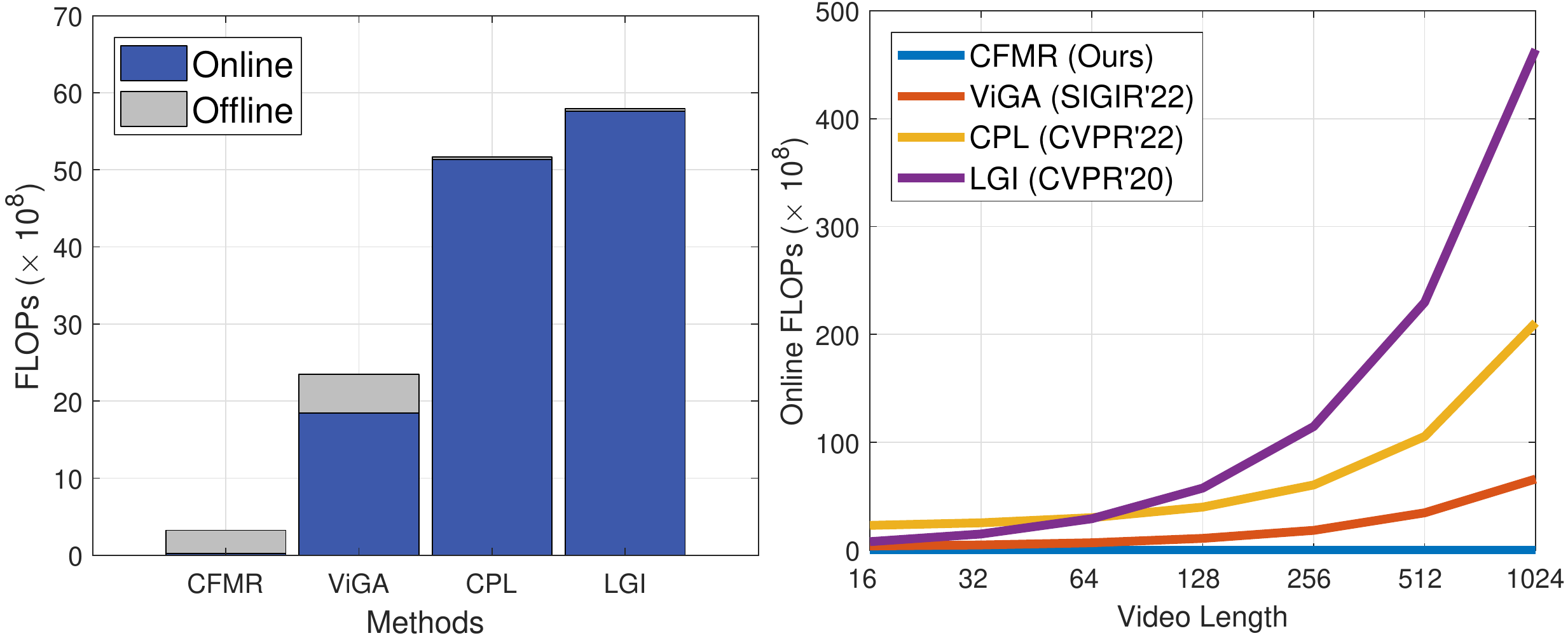}
  \caption{Comparisons w.r.t online and offline FLOPs on the ActivityNet-Captions dataset.}
  \label{fig:efficiency}
  
\end{figure}

\noindent
\textbf{Analysis on Point-guided Contrastive Learning.}
We also furtherly explore the procedure of SR module in our CFMR model. 
Concretely, we conduct experiments with different settings on the max widths $v_\text{max}$ and observe corresponding performance of moment retrieval as well as the language reconstruction. 
According to the results in Fig. \ref{fig:PCL}, we list following observations: 
(1) The reconstruction accuracy shows positive correlations to the performance of moment retrieval. 
We can see that the ablated models with higher VMR recall rates obtain more high-quality semantic reconstruction. 
It proves again the SR module is effective to learn the hidden semantics in video and text modalities, which helps generate relevant common concepts. 
(2) It is helpful to increase the scales of the temporal receptive field in PCL. 
By introducing larger max widths of positive anchors $v_\text{max}$, the CFMR achieves remarkable improvements in the performance of retrieval and language reconstruction both. 
However, we also note that such a strategy is limited as the improvements become less significant at $v_\text{max}=0.8$. 
The reason is oversized temporal receptive fields bring massive irrelevant content, which leads to degeneration in the contrastive learning among positive and negative samples. 

\begin{figure}[htb!]  
  \centering
  \includegraphics[width=\linewidth]{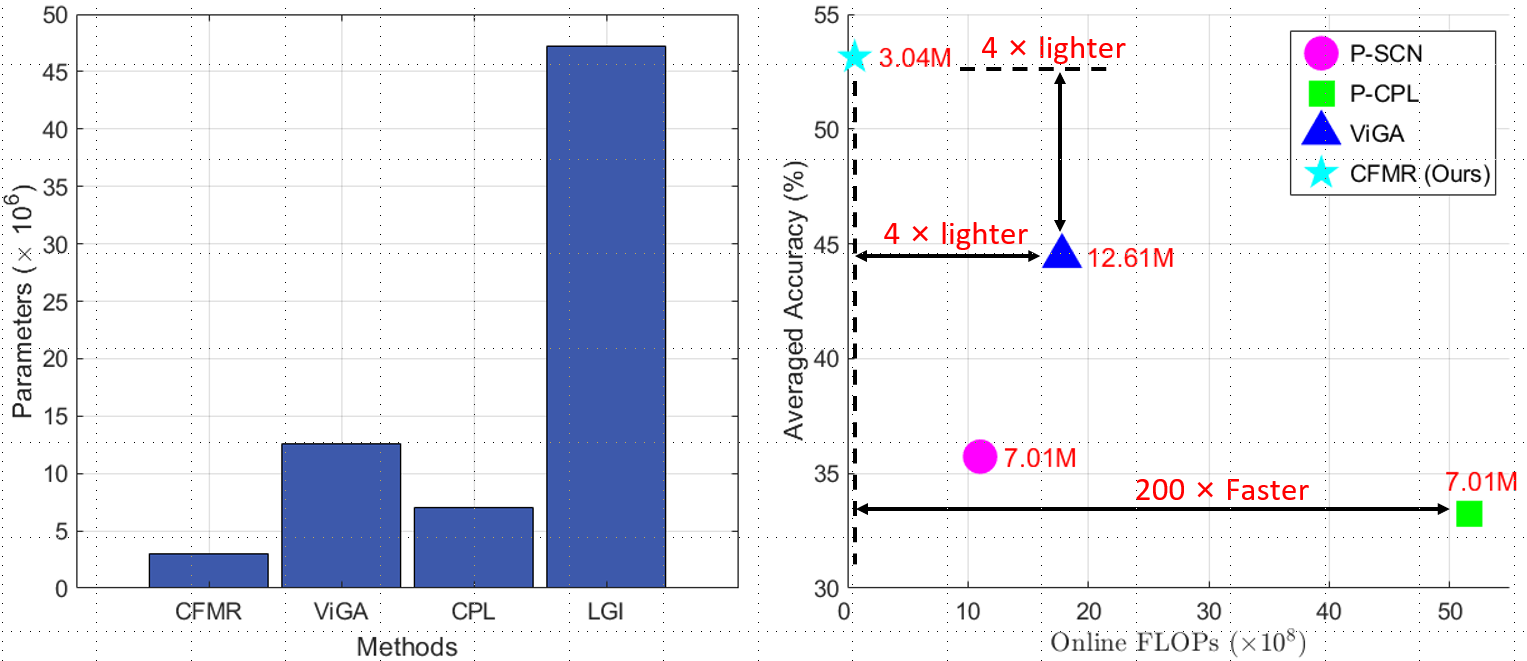}
  \caption{Analysis with respect to model parameters and comprehensive performance, including averaged accuracy (R1@IoU=0.5 and R5@IoU=0.5), model parameters ($\times 10^{6}$), and FLOPs $\times 10^{8}$ on the ActivityNet-Captions dataset.}
  \label{fig:parameters}
  
\end{figure}

\noindent
\textbf{Analysis on Efficiency.} 
We also show more details of computational consumption in the CFMR method and make a comparison with other VMR methods \cite{mun_LGI_CVPR2020, Zheng_CPL_CVPR2022, Cui_ViGA_SIGIR2022}. 
Specifically, We count the offline (processing one video) and online (processing one query and cross-modal interactions) FLOPs to furtherly demonstrate the superiority of our model.
According to the comparative results in Fig. \ref{fig:efficiency}, we obtain following observations: 
(1) The CFMR significantly reduces the proportion of online computational cost. 
Facilitated with concept-based retrieval rather than cross-modal interactions, it requires no extra online computation but only encoding users' text queries online. 
(2) With the untrimmed videos get longer, our method keeps high efficiency while the computational consumption of others increase dramatically. 
It demonstrates the previous methods heavily suffer from the complex online cross-modal interactions which are sensitive to videos. 
Contrastively, our CFMR effectively tackles this problem by eliminating this inflexible design to maintain remarkable performance.

To furtherly show the superiority of the model efficiency, we also count parameters of the CFMR method and make comparisons with the other methods. 
As is illustrated in Fig. \ref{fig:parameters}, our method is much lighter, which only requires less than one quarter of parameters of the previous PS-VMR ViGA method. 
Moreover, we comprehensively compare our CFMR method to several PS-VMR methods in Fig. \ref{fig:parameters} by observing their online FLOPs, parameters, and accuracy. 
It can be seen our CFMR achieve remarkable improvement on the averaged moment retrieval accuracy, while is 80 $\times$ faster and 4 $\times$ lighter than the existing state-of-the-art ViGA method.

\begin{table}[htb!]
  \small
  \centering
  \caption{Comparisons on model performance on the ActivityNet-CD-OOD debiased datasets with C3D features.}
  
  \resizebox{\linewidth}{!}{
      \begin{tabular}{c|cc|cc}
     \toprule
      \multirow{2}{*}{Method} & R@1,    & R@1,    & R@5,    & R@5  \\ 
      ~ & IoU=0.5 & IoU=0.7 & IoU=0.5 & IoU=0.7\\ 
      \hline

      CNM \cite{Zheng_CNM_AAAI2022} & 7.13 & 0.43 &  - & -  \\
      
      CPL \cite{Zheng_CPL_CVPR2022}& 8.71 & 1.63 & 14.24 & 3.24  \\
      
      P-CPL \cite{Zheng_CPL_CVPR2022}& 9.21 & 1.33 & 11.44 & 2.21  \\
      
      ViGA \cite{Cui_ViGA_SIGIR2022}& 20.28 & 8.12 & 30.21 & 16.13  \\
      
      \textbf{CFMR (Ours)} & \textbf{21.59} & \textbf{9.76} & \textbf{49.43} & \textbf{32.24}  \\
      \bottomrule
      \end{tabular}
  }
  \label{tab:ood}
\end{table}

\noindent
\textbf{Performance on Out-Of-Distribution Settings.} 
Furthermore, we also conduct more experiments under the out-of-distribution (OOD) settings to evaluate the robustness and generalizing strength of our CFMR method. 
Specifically, we adopt the data split proposed in \cite{Yuan_OOD_arxiv2021}, where the ground-truth moment distribution is different in the training and test splits. 
Similarly, only point-level annotations are provided in the training split to reduce the training data cost. 
We report the experimental results on the ActivityNet-CD-OOD datasets in Table \ref{tab:ood} and compared with recent FS-VMR, WS-VMR, and PS-VMR methods. 
It can be observe that our method still achieves remarkable improvements compared with the counterpart ViGA method \cite{Cui_ViGA_SIGIR2022}, especially on the metrics of R@5. 
This is because our method deploys the CMA module and PCL module, which effectively extract hidden semantics and conduct concept-based multimodal alignments with point-level supervision signals.

\begin{figure}[thb!] 
  \centering
  \includegraphics[width=\linewidth]{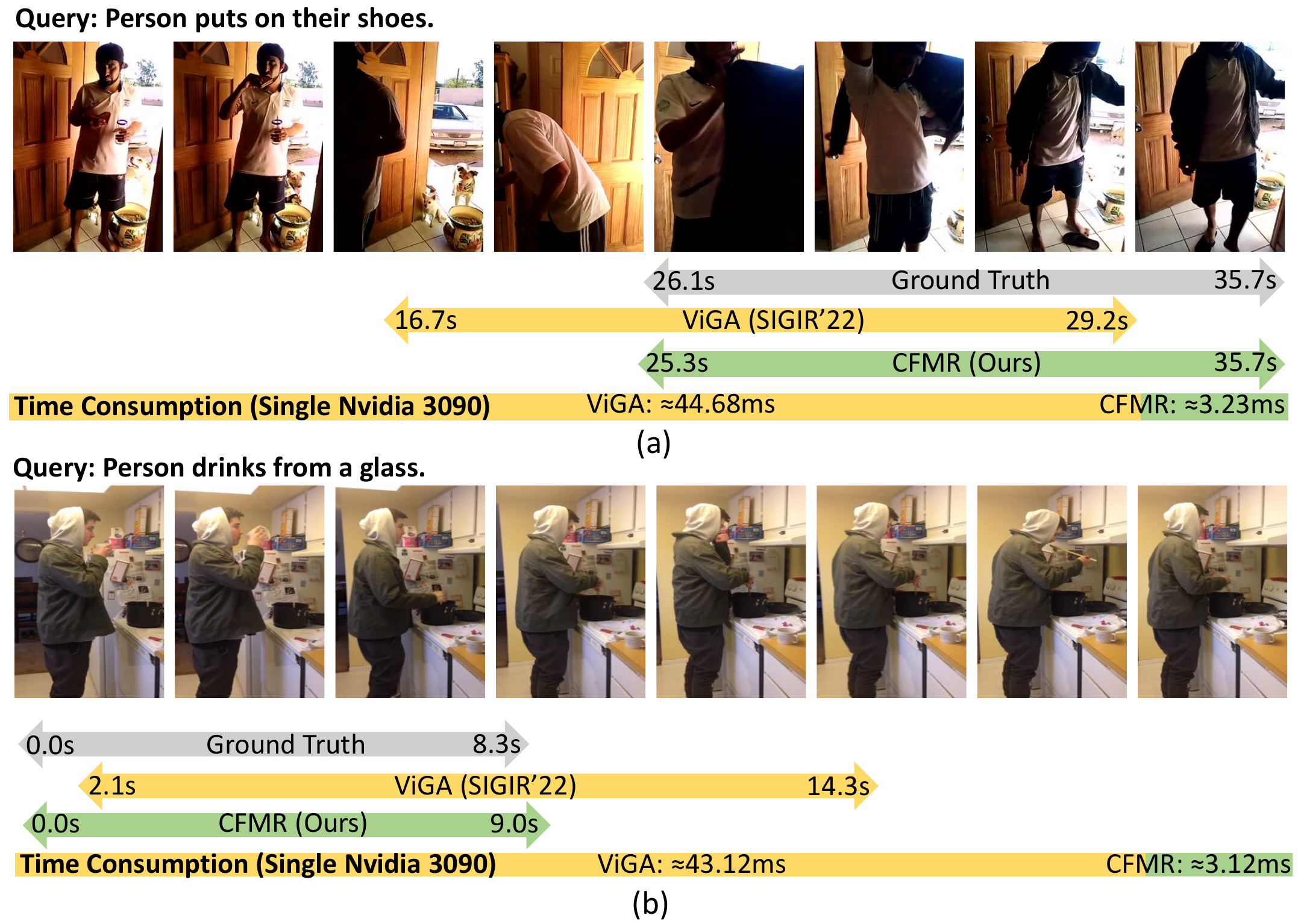}
  \caption{Visualization of two examples of moment retrieval results and time consumption on the Charades-STA dataset. }
  \label{fig:qa}
  
\end{figure}
\noindent
\textbf{Qualitative Analysis.} 
We also illustrate the qualitative analysis in Fig. \ref{fig:qa} obtained by our CFMR method and the latest counterpart ViGA method \cite{Cui_ViGA_SIGIR2022}. 
According to the visualization results, our CFMR reveals remarkable superiority in retrieval accuracy and also achieves higher retrieval efficiency over the ViGA method. 
Specifically, it takes only 3 milliseconds nearly to retrieve the most correlated moment precisely for our CFMR method, while more than 40 milliseconds for the previous ViGA method. 
Such a result explicates again the effectiveness of our proposed CFMR method, which eliminates the cross-modal interactions in the test stage and fast aligns two modalities with diverse semantic concepts to retrieve the target moment.

\section{Conclusion}
\label{sec:conc}
In this paper, we proposed a Cheaper and Faster Moment Retrieval (CFMR) method to achieve a fair trade-off among retrieval accuracy, efficiency, and annotation cost for the VMR task. 
It can be trained with much cheaper point-level supervision but shows competitive retrieval performance.
Moreover, unlike existing methods, our CFMR method bypasses the cross-modal interaction modules in previous VMR methods thus greatly optimizing model efficiency. 
Extensive experiments on three widely adopted benchmarks demonstrated the remarkable performance of the proposed CFMR method. 
For future work, we will explore more practical VMR methods to furtherly facilitate the development of relevant applications.


\bibliographystyle{ACM-Reference-Format}
\bibliography{ref}










\end{document}